\documentclass[5p,twocolumn]{elsarticle}

\usepackage{amsmath,amssymb} 
\usepackage{color}
\usepackage{amsmath,mathtools}
\usepackage[table]{xcolor}
\usepackage{multirow}
\usepackage{subcaption}
\usepackage{tabularx}
\usepackage{algorithm}
\usepackage{algpseudocode}

\usepackage{url}

\begin{document}  
	
	\begin{frontmatter}
		\title{Evaluation of the Spatio-Temporal features and GAN for Micro-expression Recognition System}
		
		\author[add1]{Sze-Teng Liong} 
		\ead{stliong@fcu.edu.tw}
		\author[add2]{Y.S. Gan}
		\ead{ysgn88@gmail.com}
		\author[add3]{Danna Zheng} 
		
		\author[add3]{Shu-Meng Li} 
		
		\author[add3]{Hao-Xuan Xu}
		
		\author[add3]{Han-Zhe Zhang}
		
		\author[add3]{Ran-Ke Lyu}
		
		\author[add4]{Kun-Hong Liu\corref{cor1}}
		\ead{lkhqz@xmu.edu.cn}

		\cortext[cor1]{Corresponding author}
		\address[add1]{Department of Electronic Engineering, Feng Chia University, Taichung, Taiwan}
		\address[add2]{Research Center for Healthcare Industry Innovation, NTUNHS, Taipei, Taiwan} 
		\address[add3]{School of Electrical and Computing Engineering, Xiamen University Malaysia, Jalan Sunsuria, Sepang, Selangor, Malaysia} 
		\address[add4]{School of Software, Xiamen University, Xiamen, China}

		\begin{abstract}
			Owing to the development and advancement of artificial intelligence, numerous works were established in the human facial expression recognition system.
			Meanwhile, the detection and classification of micro-expressions are attracting attentions from various research communities in the recent few years.
			In this paper, we first review the processes of a conventional optical-flow-based recognition system, which comprised of facial landmarks annotations, optical flow guided images computation, features extraction and emotion class categorization. 
			Secondly, a few approaches have been proposed to improve the feature extraction part, such as exploiting GAN to generate more image samples. 
			Particularly, several variations of optical flow are computed in order to generate optimal images to lead to high recognition accuracies. 
			Next, GAN, a combination of Generator and Discriminator, is utilized to generate new ``fake'' images to increase the sample size.
			Thirdly, a modified state-of-the-art convolutional neural networks is proposed.
			To verify the effectiveness of the the proposed method, the results are evaluated on spontaneous micro-expression databases, namely SMIC, CASME II and SAMM. 
			Both the F1-score and accuracy performance metrics are reported in this paper.

		\end{abstract}
		
		
	\end{frontmatter}

\section{Introduction}
\label{sec:introduction}

Micro-expression (ME) is a kind of natural human expression that normally occurs when a person tries to conceal his or her genuine emotion. 
Different from other genuine expressions, MEs are the tiny movements of facial muscles and usually last for half a second~\cite{yan2013fast}. 
Due to its uncontrollable and involuntary characteristics, MEs always show the real feelings of a person.
Thus, recognizing ME helps the researchers especially in the psychology aspect to discover hidden insights.
Besides, MEs recognition are applied to several fields, such as the police interrogation~\cite{vrij2005police},  psychological clinical diagnosis~\cite{endres2009micro}, social interaction~\cite{lodder2016loneliness}, judicial system~\cite{qian2009grand}, political elections~\cite{stewart2009presidential} and national security~\cite{holmes2011national}.
However, due to the subtlety and rapid atrributes, it poses a great challenge for normal people to notice the occurrence of ME from their naked eyes in the real-time conversations. 
Similar to the normal facial expression, a.k.a., macro-expression, it can be classified into six basic emotions: happy, sad, surprise, fear, anger and disgust. 
In brief, ME is first discovered in 1969, when Ekman~\cite{ekman1969nonverbal} analyzed an interview video of a psychotherapist and a patient who attempted to commit suicide. 
When Ekman watched the video frame by frame, he found out that the patient tried hide her sad feeling by covering up with smile during the interview process.
Later, Ekman and Friesen ~\cite{ekman1978facial} created a model called Facial Action Coding System (FACS) to identify each action units (AUs).
Action unit describes the facial muscle motions in certain direction.

Classically, an automatic ME recognition system is divided into two steps: ME spotting and ME recognition. 
The former is to identify the ME interval or some specific important frames; while the latter is to perform the emotion class classification. 
Currently, there are many methods proposed in literatures to perform the ME spotting and recognition tasks.
Thus far, the highest recognition accuracy achieved is 78.14\%, which is conducted by Li et al.~\cite{li2018towards} using the HIGO method and SVM classifier.
Hence, there is still room for improvement to build a better system. 

\section{Related Work}
This section reviews the micro-expression databases, discusses the techniques used in ME spotting and recognition, and analyzes the problems faced in the previous works.

\subsection{Database}
To date, there are six popular ME databases established for the researchers especially from computer vision field for algorithm development and analysis. 
The databases included USF-HD~\cite{shreve2011macro}, SMIC~\cite{pfister2011recognising,li2013spontaneous}, CASME~\cite{yan2013casme}, CASME II~\cite{yan2014casme}, CAS(ME)$^2$~\cite{qu2018cas} and SAMM~\cite{davison2018samm}. 
The brief information of these six databases are shown in Table~\ref{table:db}.
Among the six databases, CASME II has the most abundant samples that contains 247 videos and it is commonly served as the baseline database for algorithm evaluation.
It is observed that the database are having small data size.
This is because some emotions are more likely to trigger, resulting in uneven distribution of samples. 
On the other hand, the ethnic and geographical distribution of the participants in each database are different, hindering the development of feature extractors methods and it is also difficult for data generalization.

\setlength{\tabcolsep}{5pt}
\begin{table*}[tb]
	\begin{center}
		\caption{Available Micro-expression Databases}
		\label{table:db}
		\begin{tabular}{lcccccc}
			\noalign{\smallskip}
			\hline
			\noalign{\smallskip}
			Database
			& Year
			& \# Participant 
			& \# Video
			& Frame Rate
			& Resolution
			& \# Emotion\\
			\hline
			\noalign{\smallskip}
			USF-HD 
			& -
			& - 
			& 100
			& 30
			& 170$\times$1280
			& 6\\
			
			\noalign{\smallskip}
			SMIC
			& 2013
			& 16
			& 164
			& 100
			& 640 $\times$ 480
			& 3 \\ 
			
			\noalign{\smallskip}
			CASME
			& 2013
			& 35
			& 195
			& 60
			& 640 $\times$ 480
			& 7 \\
			
			\noalign{\smallskip}
			CASME II
			& 2014
			& 35
			& 247
			& 200
			& 640 $\times$ 480
			& 5 \\
			
			\noalign{\smallskip}
			CAS(ME)$^2$
			& 2018
			& 22
			& 57
			& 30
			& 640 $\times$ 480
			& 4 \\
			
			\noalign{\smallskip}
			SAMM
			& 2018
			& 32
			& 159
			& 200
			& 2040 $\times$ 1088
			& 7 \\

			\hline
			
		\end{tabular}
	\end{center}
\end{table*}

\subsection{Micro-Expression System}
An ME micro-expression system comprised of three main basic stages, which are pre-processing, feature extraction and classification namely.
The pre-processing techniques include the face detection and registration, face masking and regions of interest acquisition, important frames selection.
Most of the research works focus on improving the feature extractors, compared to the pre-processing and classification stages.
Feature extraction is to represent an image with minimum feature dimension whilst retaining meaningful ME details.
The feature extractors can be categorized into handcrafted and deep learning methods.
The following subsections elaborate the state-of-the-arts of these three stages.

\subsubsection{Pre-processing}
\begin{figure}[t!]
	\centering
	\includegraphics[width=1\linewidth]{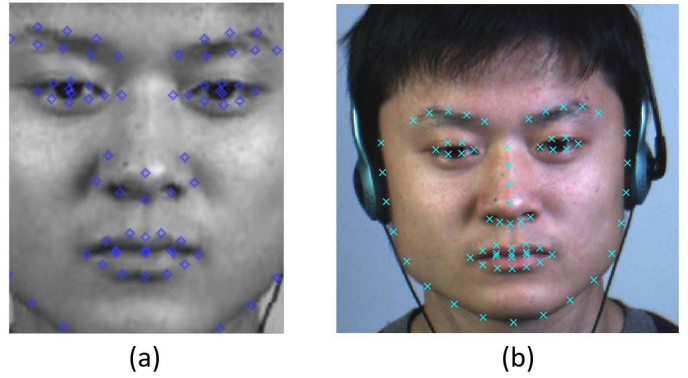}
	\caption{Example of locating the landmark coordinates using: (a) Face++, and (b) DRMF}
	\label{fig:landmark68}
\end{figure}

\paragraph{Face detection and registration}
To detect the position of the face, the landmarks of the face are first located. 
The example of the face with landmark coordinates annotated is shown in Figure~\ref{fig:landmark68}~\cite{liong2017micro}.
It can be seen that there is a total of 68 labeled locations on the face.
Prototypicality, there are three techniques commonly used in locating the facial landmark points, namely: Active Shape Model (ASM)~\cite{cootes1995active}, Active Appearance Model (AAM)~\cite{edwards1998face} and Discriminative Response Map Fitting (DRMF)~\cite{asthana2013robust}.

ASM is an algorithm based on the Point Distribution Model (PDM).
Basically, it first selects the landmarks from a training set manually.
Then, a feature-based method, Procrustes\cite{xia2016spontaneous}, is employed for face alignment and registration.
It constructs the local features for each landmark to iteratively search new landmarks and obtains the shape model.
An advantage of the model is that it has significant constraints on the contour shape. 
However, its iteratively search strategy limits the computational efficiency. 
AAM model is an improved version of ASM by adding the texture features from the entire face area. 
On the other hand, DRMF is based on the Cascade Pose Regression ~\cite{dollar2010cascaded} that uses the Support Vector Regression (SVR) ~\cite{drucker1997support} to model a regression function.
It adopts the shape-dependent Histogram of Gradient (HOG)~\cite{zhu2006fast} features as the input to cascade the predicted face shape.
There are a few online tools such as Face++~\cite{face2013} and Openface~\cite{amos2016openface}, which are executed in real-time performance without installing any specialist hardware.

In 2013, Sun et al.~\cite{sun2013deep} applied convolutional neural network (CNN) for face landmark detection and proposed a cascaded CNN (with three levels).
Since the CNN has superior abilities of feature extraction, it is more precise and robust in the landmark detection and resolve the local optimal issues.
However, it fails to deal with the occlusion problems.
On the other hand, Huang et al.~\cite{huang2015coarse} introduced a algorithm from coarse to fine to divide the face landmarks into internal landmarks and contour landmarks, using the Face++, and it leads to high-precision positioning.
In 2016, Zhang et al.~\cite{zhang2016joint} presented a multi-task Cascaded Convolutional Networks (MTCNN) to simultaneously conduct face detection and face landmarks detection. 
Concisely, it consists of three cascaded multitasking CNN: Proposal Network (P-Net), Refine Network (R-Net), and Output Network (O-Net).
A year later, Kowalski et al.~\cite{kowalski2017deep} proposed a new cascaded deep neural network, Deep Alignment Network (DAN). 
By adding a landmarks heat map as a supplement, DAN extracts the features from the whole picture to obtain more accurate landmark positioning.
However, there is a significant disadvantage in these three methods, which is a high computationally complexity due to the complicatedness of the mathematical derived operations.

For the face registration, there exists two approaches: area-based method and feature-based method.
The former method is based on the correspondence of the entire image, such as 2D-DFT\cite{davison2018objective}, while the latter is based on the correspondence of regions, points, and line features (i.e., affine transform, Local Weighted Mean\cite{goshtasby1988image} and Procrustes analysis).

\paragraph{Face Masking and Region Partitioning}
Face masking is to reduce the effects of irrelevant factors that do not correspond to the desired facial movements. 
To eliminate the noise generated from the eye blinking motions,  Shreve et al.~\cite{shreve2011macro} and Liong et al.~\cite{liong2016automatic} proposed to mask the eye regions for each image.
In addition,~\cite{shreve2011macro} covers other regions such as nose and mouth with a T-shaped mask.
Davison et al.~\cite{davison2018objective} identified 26 FACS-based facial regions heuristically to faciliate the facial feature localization.
Moilanen et al.~\cite{moilanen2014spotting}, Davison et al. \cite{davison2015micro,davison2018samm}, Wang et al.~\cite{wang2016main}, Liong et al.~\cite{liong2014subtle} and Li et al.~\cite{li2018towards} divided the entire face into several equal parts.
On the other hand, Polikovsky et al.~\cite{polikovsky2009facial,polikovsky2013facial}, Liong et al.~\cite{liong2015automatic,liong2016automatic,liong2018less,liong2018hybrid}, Davison et al.,~\cite{davison2018objective} and Li et al.~\cite{li2018ltp} selected certain region-of-interests (RoIs) from the face.
Generally, the method with RoIs partitioning yields to promising performance as it is better to represent the local feature efficiently.

\begin{figure*}[t!]
	\centering
	\includegraphics[width=0.9\linewidth]{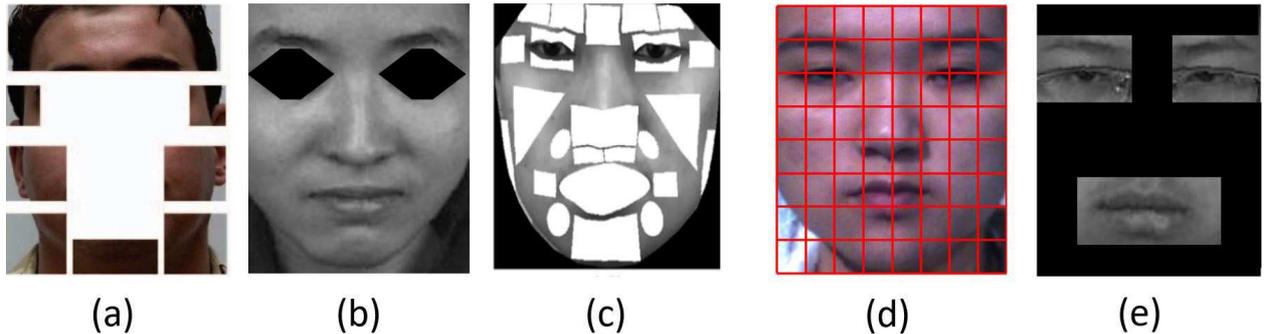}
	\caption{Example of the face masking and partitioning: 
		(a) A T-shaped mask is applied to obtain the eight important regions~\cite{shreve2011macro};
		(b) The eyes are masked to avoid eye blinking~\cite{liong2016automatic};
		(c) The face is divided into 26 regions~\cite{davison2018objective};
		(d) The face is partitioned into several equal parts~\cite{liong2014subtle};
		(e) Only the eyes and mouth regions are considered~\cite{shreve2014automatic}}
	
	\label{fig:face_masking}
\end{figure*}

\paragraph{Frames selection and Magnification}
Since the MEs are very subtle, most of the sample videos are elicited using a high frame rate camera.
Repetitive frames with very little changes in movements might result in data redundancy.
In other words, not all the frames are important.
A frame selection step dynamically choose the significant frame that contain meaningful expression details.
For instance, Temporal interpolation method (TIM)~\cite{zhou2011towards,pfister2011recognising,wang2014micro,li2018towards} is often used in the previous experiments to standardize the frame length for all videos.
On another note, Le Ngo et al.~\cite{le2017sparsity} show that the recognition accuracy is improved by applying Sparsity-Promoting Dynamic Mode Decomposition (DMDSP).
They demonstrate that DMDSP is effective when interpolating small number of frames, but larger number of frames degrades the recognition performance because of over interpolation. 
To enlarge the subtle motion changes of MEs, motions magnification operation can be applied, such as the Eulerian Video Magnification (EVM)~\cite{wu2012eulerian}.
An improved version of EVM, namely, Global Lagrangian Motion Magnification (GLMM), is recently utilized by Le Ngo et al.~\cite{le2018micro}, whereby they report that higher ME recognition accuracy results can be achieved. 

\subsubsection{Feature Extraction}
\paragraph{Handcrafted Approach}

Many pioneer ME research works~\cite{pfister2011recognising,li2013spontaneous,yan2013casme,yan2014casme,qu2018cas,davison2018samm} applied Local Binary pattern (LBP) variants to evaluate the videos collected.
LBP is a feature descriptor that effectively represents a two-dimensional gray-scale image in a compact binarized vector.
The LBP family includes LBP~\cite{ojala1996comparative}, Local Binary Pattern on Three Orthogonal Planes (LBP-TOP)~\cite{pfister2011recognising,zhao2007dynamic,house2015preprocessing}, Local Binary Pattern with Six Intersection Points (LBP-SIP)~\cite{wang2014lbp}, Local Binary Pattern with Mean Orthogonal Planes (LBP-MOP)~\cite{wang2015mop},
Discriminative Spatiotemporal Local Binary Pattern with Revised Integral Projection (DiSTLBP-RIP)~\cite{xiaohua2017discriminative} and Spatiotemporal Completed Local Quantization Patterns (STCLQP)~\cite{huang2016spontaneous}.
LBP is insensitive to illumination change, computational simplicity, capable of handling a variety of spatial information, robust to rotation and translation.
LBP-TOP extracts features from both the spatial and temporal perspectives, for instance, from the three planes (XY, XT, YT planes).
To reduce some of the duplicated computation in LBP-TOP, LBP-SIP is introduced to consider six neighboring unique points for each centre pixel.
Thus, it produces 2.4 times lesser feature length per video compared to LBP-TOP yet with higher recognition on SMIC and CASME II datasets.
LBP-MOP computes three mean images for each video and it outperforms LBP-TOP on CASME II dataset. 
Notably, the average time for extracting the features in each video using LBP-MOP is 38 times faster than that of LBP-TOP.  
On the other hand, DiSTLBP-RIP is a combination of LBP with integral projection and a comparable accuracy is achieved with simpler computation.
STCLQP is mathematically more complex than LBP-TOP, because it collects more information which are sign, magnitude and orientation components. 
Wang et al.~\cite{wang2014microtics} extracted LBP-TOP features with Tensor Independent Color Space (TICS) and discovered that the recognition performance in TICS is better than that in both RGB and gray-scale.
Generally, LBP-based methods are widely used, but it does not reflect the motion changes from certain face muscles intuitively.

Aside from LBP family, there is an optical flow family which are commonly utilized in ME recognition system.
Basically, optical flow describes the apparent motion of the facial muscle movements based on the brightness patterns.
The example of the optical flow family consists of Optical Strain Feature (OSF)~\cite{liong2014subtle}, Optical Strain Weight (OSW)~\cite{liong2014subtle}, Fuzzy Histogram of Oriented Optical Flow (FHOOF)~\cite{happy2017fuzzy}, Fuzzy Histogram of Optical Flow Orientations (FHOFO)~\cite{happy2017fuzzy}, Bi-weighted Oriented Optical Flow (Bi-WOOF)~\cite{liong2018less}, Facial Dynamics Map (FDM)~\cite{xu2017microexpression} and Main Directional Mean Optical flow (MDMO)~\cite{liu2016main}.
Particularly, FHOOF, FHOFO, MDMO are insensitive to illumination changes and have better performance than LBP-TOP and HOOF. 
However, the weight assigned to FHOOF is highly dependent on motion magnitudes of various MEs, yet, FHOFO overcomes this drawback. 
In~\cite{liong2018less}, Liong et al. proposed to use Bi-WOOF to extract the significant features and demonstrated that using onset and apex frames to represent the entire video sequence is feasible.

On the other hand, there is a gradient-based family which computes changes of the adjacent frames and produce the output in both the  horizontal and vertical directions.
The gradient-based features includes 3D-gradient~\cite{polikovsky2009facial}, Histograms of Oriented Gradients (HOG)~\cite{dalal2005histograms} and Histogram of Image Gradient Orientation (HIGO)~\cite{li2018towards}.
HIGO is simpler and less susceptible to illumination changes compared to HOG. 
~\cite{li2018towards} shows that both the HOG and HIGO outperform LBP-TOP in CASME II dataset.
Gradient-based features evolved from holistic-based to local-based.
They have similar limitations with LBP-based family: the pixel-level attention. 

\paragraph{Deep Learning Approach}
Deep learning architecture has attracted much attention from the computer vision communities due to its viability in a broad range of applications.
Concretely, for the ME recognition system, Peng et al.~\cite{peng2017dual} suggest a Dual Temporal Scale Convolution Neural Network (DTSCNN) that comprised of two streams CNN to accommodate the input data with different frame rate. 
The results show that DTSCNN outperforms STCLQP, MDMO, and FDM by approximately 10\% in accuracy performance metric. 
Kim et al.~\cite{kim2017multi} proposed to combine CNN with long-term memory (LSTM) and remarkable results are obtained.
It has been demonstrated that shallow CNN architectures are practically workable in the ME recognition system~\cite{liong2018off,liong2019shallow, gan2018bi, zhang2018smeconvnet} .
Overall, although the transfer learning~\cite{patel2016selective} and data augmentation~\cite{takalkar2017image} are performed, the limited number and imbalance emotion class distribution issues pose a great challenge to develop robust feature extractors.

\subsubsection{Classification}
Classification stage is to categorize the emotion of a video according to the feature extracted. 
Support Vector Machine (SVM)~\cite{gunn1998support} is the most popular classifier adopted in ME recognition system.
Compared to other classifiers such as Random Forest (RF)~\cite{liaw2002classification}, sparse representation classifier (SRC)~\cite{yang2012beyond} and Relaxed K-SVD~\cite{zheng2016relaxed}, SVM appears to be more consistent across all the databases with distinct features extracted.
Due to the emergence of deep learning, the Softmax classifier, normally served as the final fully connected layer, has been employed in the recent works~\cite{kim2017multi, gan2019off,gan2018bi, liong2019shallow}.
Its outstanding discriminative characteristics is beneficial, especially in dealing the high-level features.

\section{Methodology}

\begin{figure*}[thpb]
	\centering
	\includegraphics[width=1\linewidth]{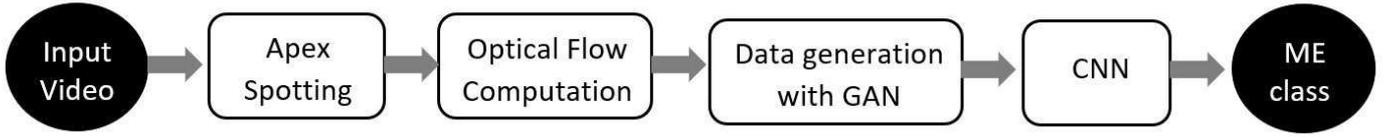}
	\caption{Flow diagram of the proposed methodology}
	\label{fig:flowchart}
\end{figure*}

This section discusses the methods used for the whole process of micro-expression recognition.
The flowchart of the proposed approach is illustrated in Figure~\ref{fig:flowchart}.
There are three primary objectives in this paper:
\begin{enumerate}
	\item An analysis of both the sparse and dense optical flow methods and discover the effect towards the ME recognition system.
	\item To increase the data sample size, several Generative Adversarial Network (GAN) approaches are exploited to generate ``fake'' optical flows images. 
	\item Implementation of a state-of-the-art CNN architecture and modification of the structure for feature enhancement and classification operations.
\end{enumerate}

Details of each step are explained in the following subsections.

\subsection{Apex frame spotting}
\label{subsec:spot}
Inspired by~\cite{liong2018less, liong2018off}, only two frames, viz., the onset and apex frame are selected to represent the whole video clip.
The main advantages of utilizing two frames per video are computational simplicity, high efficiency and redundancy elimination.
Apex frame is the instant indicating of the most expressive emotional state in a video.
Theoretically, the apex frame portrays the highest facial muscle movement changes compared to the onset frame.
Onset frame is the moment denoting the beginning of an emotion.
To spot the apex frame, we directly adopt the Divide \& Conquer (D\&C) method proposed by Liong et al.~\cite{liong2018less}.
Concretely, the algorithm consists five steps:
1) Multiple local peaks of whole video clips are determined; 
2) Frame sequence is divided into two halves equally (i.e., a 40 frames video sequence is split into two sub sequences containing frame 1-20 and 21-40);
3) Magnitude of the detected peaks are summed up for each of the sub sequence; 
4) The sub sequence with the higher magnitude will be remained and the other sub sequence will be neglected;
5) Step 2 to 4 are repeated until the single final frame is identified.

\subsection{Optical Flow Computation}

Optical flow describes the dynamic changes of an object and represent the movements using a two-dimension vector field (i.e., horizontal and vertical optical flows).
There are several optical flow methods in the literature, namely, Robust Local Optical Flow (RLOF)~\cite{senst2012robust}, Horn \& Schunck~\cite{horn1981determining}, TVL1~\cite{wedel2008duality}, Farneback~\cite{farneback2003two}, LiteFlowNet~\cite{hui2018liteflownet}, PWC-Net+~\cite{sun2018pwc}, FlowNet~\cite{dosovitskiy2015flownet} and Lucas Kanade~\cite{barron1992performance}. 
Among the optical flow methods, we exploit Farneback, Horn~\&~Schunck, Lucas Kanade, TVL1 and RLOF to compute the flow fields of the onset and apex frames for all the video sequence.
The horizontal and vertical optical flow are expressed as ($p,q$).
There are three other optical flow properties that can be derived from ($p,q$): magnitude, orientation and optical strain. 
Succinctly, to obtain the magnitude and orientation, the horizontal and vertical flow components,  $\vec{o} = (p, q)$, are converted from Euclidean coordinates to polar coordinates:

\begin{equation} \label{eq:rho}
\rho_{x,y} = \sqrt{p^2_{x,y} + q^2_{x,y}},
\end{equation}

and 
\begin{equation} \label{eq:theta}
\theta_{x,y} = tan^{-1}\frac{q_{x,y}}{p_{x,y}},
\end{equation}

\noindent where $\rho_{x,y}$ and $\theta_{x,y}$ are the magnitude and orientation respectively.

In terms of displacements, the optical strain ($\varepsilon$) is expressed as: 

\begin{equation} \label{eq:tensor}
\varepsilon = \frac{1}{2} [\nabla \bf u + (\nabla \bf u)^{\it T} ]
\end{equation}

\noindent or can be formulated as:

\begin{equation}
\varepsilon = \begin{bmatrix}
\varepsilon_{xx} = \frac{\partial u}{\partial x} & \varepsilon_{xy} = \frac{1}{2}(\frac{\partial u}{\partial y} + \frac{\partial v}{\partial x}) \\[1em]
\varepsilon_{yx} = \frac{1}{2}(\frac{\partial v}{\partial x} + \frac{\partial u}{\partial y}) & \varepsilon_{yy} = \frac{\partial v}{\partial y}
\end{bmatrix}
\end{equation}

\noindent where the diagonal strain components, ($\varepsilon_{xx}, \varepsilon_{yy}$) are normal strain components and ($\varepsilon_{xy}, \varepsilon_{yx}$) are shear strain components. 
Normal strain measures the changes in length along a specific direction, whereas shear strains measure the changes in two angular directions.


\subsection{Data Generation with GAN}

To evaluate the robustness of a feature extractor,~\cite{liong2019shallow} combines the video samples from three databases with different nature.
Specifically, CASME II, SMIC and SAMM are combined into a composite database, therefore, it contains 442 videos and three emotion classes (i.e., positive, negative and surprise) from 68 participants.
To further increase the size of input sample to the classifier, a neural network, namely, GAN is utilized to automatically generate new optical flow images.
Particularly, two GAN network structures are experimented: (1) Auxiliary Classifier GAN (AC-GAN) ~\cite{odena2017conditional}, and; (2) Self-Attention GAN (SAGAN) ~\cite{zhang2018self}.

The main idea of GAN is from a two-player game, where one acts as the generator model, the another one is served as the discriminator model.
In brief, a random noise $z$ is first input into the generator model and output an image $A$. 
Then, the real image $B$ together with the image $A$ is passed to the discriminator, it produces the probability contributions over its truth degree and over the real image.
The aim of the generator is to generate an image as real as possible, and the discriminator attempts to distinguish the correctness of the real and fake generated images.
After several repetitions of training processes, an discriminator outputs probability of 0.5 would be expected to indicate the balance of the model.
The pseudocode of the GAN algorithm is listed in Algorithm~\ref{alg:alg}.
In each of the game, the discriminator produces either 1 or 0 by function $g$:
\begin{equation}
\label{eq:1_0}
g(x) = \begin{dcases*}
1,  & x = Real Image\\
0, & x = Fake Image
\end{dcases*}
\end{equation}
Note that, throughout the GAN learning process, back  propagation is employed to reduce the training losses by optimizing the weights of the neurons.

\begin{algorithm*}
	\caption{Generative Adversarial Networks}
	\label{alg:alg}
	
	\begin{algorithmic}[1]
		\State$k \gets$ step
		\State$K \gets$ number of training iterations
		\Repeat 
		
		\Repeat 
		\State{Sample $m$ noise samples $\{z^{(1)}, \dots, z^{(m)} \}$ from noise prior $\delta_g(z)$ }
		\State{Sample $m$ examples $\{x^{(1)}, \dots, x^{(m)} \}$ from noise prior $\delta_data(x)$ }
		\State{Update the discriminator via stochastic gradient:}
		\State{$ \nabla_{\theta d} \frac{1}{m} \sum\limits_{i=1}^{m} [\text{log } D(x^{(i)}) + \text{log } (1-D(G(z^{(i)})))]$}
		\State{}
		\Until{$k = K$} 
		
		\Repeat 
		\State{Sample $m$ noise samples $\{z^{(1)}, \dots, z^{(m)} \}$ from noise prior $\delta_g(z)$}
		\State{Update the generator via stochastic gradient:}
		\State{$ \nabla_{\theta g} \frac{1}{m} \sum\limits_{i=1}^{m} [\text{log } (1-D(G(z^{(i)})))]$}
		
		\State{}
		\Until{$k = K$} 
		
		\Until{the stopping criterion is met} 
	\end{algorithmic}
\end{algorithm*}

The brief introductions of AC-GAN and SAGAN are explained as follows:

\begin{enumerate}
	\item \textbf{Auxiliary Classifier GAN}\\
	Odena et al.~\cite{odena2017conditional} suggest an objective function where the log-likelihood of the correct source, LS, and the log-likelihood of the correct class, LC are defined as:
	
	\begin{equation}
	\begin{split}
	L_S = &E[\text{log} P(S = real| X_{real})] \\
	& +  E[\text{log} P(S = fake| X_{fake})] ,
	\end{split}
	\end{equation}
	
	\noindent and 
	
	\begin{equation}
	\begin{split}
	L_C = &E[\text{log} P(C = c| X_{real})] \\
	& +  E[\text{log} P(C = c| X_{fake})] 
	\end{split}
	\end{equation}
	The discriminator and generator are optimized by maximizing the $L_S + L_C$ and $L_C - L_S$, respectively.
	The example of the AC-GAN structure is illustrated in Figure~\ref{fig:acgan}.\\
	
	\begin{figure*}[t!]
		\centering
		\includegraphics[width=1\linewidth]{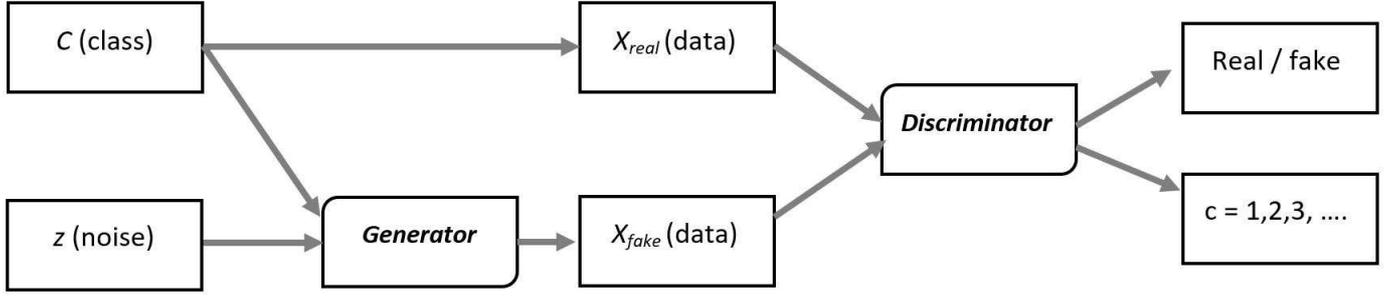}
		\caption{Auxiliary classifier GAN (AC-GAN) architectures, where x is the real image, c the class label and z the noise vector}
		\label{fig:acgan}
	\end{figure*}

	\item \textbf{Self-Attention GAN}\\
	SAGAN is a model that improved by the Natural-Language Processing (NLP) where uses attention to focus on certain important areas on the image.
	For traditional convolution GAN, the convolution operation is suitable for processing neighborhood information because of the restriction of receptive field limit. 
	However, the feature correlation of pixels that are far apart requires several convolutional layers. 
	The model is not able to capture the global information if the convolution kernel is not large enough.
	In contrast, if the kernel is considerably large, it requires more computation time and leads to low efficiency.
	SAGAN utilizes the image information from all feature locations to generate image detail, and at the meantime, ensures that the discriminator is capable to identify the consistency between two features that are far from each other. 
	In addition, a normalization operation is carried out to enhance the stability and efficiency during the model training process.
\end{enumerate}

\subsection{Convolutional neural network for Feature Classification}
\begin{figure}[thpb]
	\centering
	\includegraphics[width=1\linewidth]{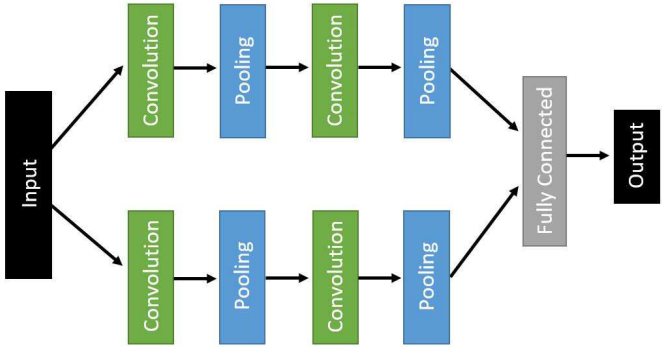}
	\caption{Example of a CNN architecture that comprised of two pairs of convolutional and pooling layers}
	\label{fig:conv_pool}
\end{figure}

A typical convolutional neural network (CNN) consists of several pairs of convolutional layer and pooling layer followed by some fully connected layers.
The convolutional and pooling layers are used for feature enhancement or feature extraction.
Concisely, each convolutional layer is made up of several kernels or filters (i.e., usually 6, 16, 32 or 64 kernels, depends on the context of the input image).
The kernels are the matrices, a.k.a., the weights, which will be multiplied to the input image. 
As a result, the original input image becomes ``thicker'' or ``deeper''.
In order to reduce the information redundancy, the common practice is to place a pooling layer after each convolutional layer. 
A pooling layer consists of certain kernel number, but here the kernels' function is to scan the inputs and identify the most representative value in a certain pixels region. 
For instance, after the applying a $2\times 2$ pooling layer kernel on a $2\times 2$ image produced by the convolutional layer, the image will be reduced to a single number. 
There exists a variety of pooling methods, such as max pooling, min pooling and stochastic pooling.
An example of a simple CNN architecture that comprised of two pairs of convolutional and pooling layers is illustrated in Figure~\ref{fig:conv_pool}.

Conventionally, after the input image passes through a few pairs of convolutional and pooling layers, one or more fully connected layer are then placed.
A fully connected layer accepts matrices as inputs, it connects every input matrix to all the existing output channels.
This fully connected layer also served as the classifier to produce corresponding emotion classes as the output. 
In order to avoid the phenomena of overfitting or underfitting in the CNN model training process, the loss function will be observed according to the difference between the predicted emotion class attained by the network and the true emotion class.
The gradient of the loss is calculated, which will be back propagated to every kernel. At the same time, the weights (i.e., in convolutional layers and fully connected layers) will be updated. 
The training process is terminated when training loss gradually decreases across the training iteration (i.e., epoch).

We tend to enhance the OFF-ApexNet~\cite{liong2018off} architecture to improve the recognition results.
The OFF-ApexNet accepts the horizontal and vertical optical flow images, then passes them to two streams of the convolutional and max pooling layers.
The outputs of both parallel layer sets are concatenated to serve as the input to the three fully connected layers. 
To compare the recognition performance of the original OFF-ApexNet architecture to the modified version, a re-implementation of original OFF-ApexNet is conducted and results are reported in the experiment section later.
In particular, the modifications made on the OFF-ApexNet are listed as follows:
\begin{enumerate}
	\item A parallel stream is added which allows to take in another type of optical flow information (i.e., magnitude, orientation, optical strain, etc.). 
	Thus, instead of only extracting the features from two directions of the optical flows ($p,q$), CNN can learn more ME details and provide more clues for the ME characteristics.
	\item Rather than concatenating the output features at the beginning of the fully connected layer, a multiplication operation is used to reduce the dimension of the features whilst maintaining the features quality.
	\item A visualization of features in different layers is implemented and the principle component analysis is conducted for the classification output.
	With these extra functions added to the model, the whole training process would seems to be more convincing due to the network transparency.
	Consequently, the mistakes made and the ambiguous behaviors during the training process are easily inspected and spotted.
\end{enumerate}

\begin{figure*}[thpb]
	\centering
	\includegraphics[width=0.7\linewidth]{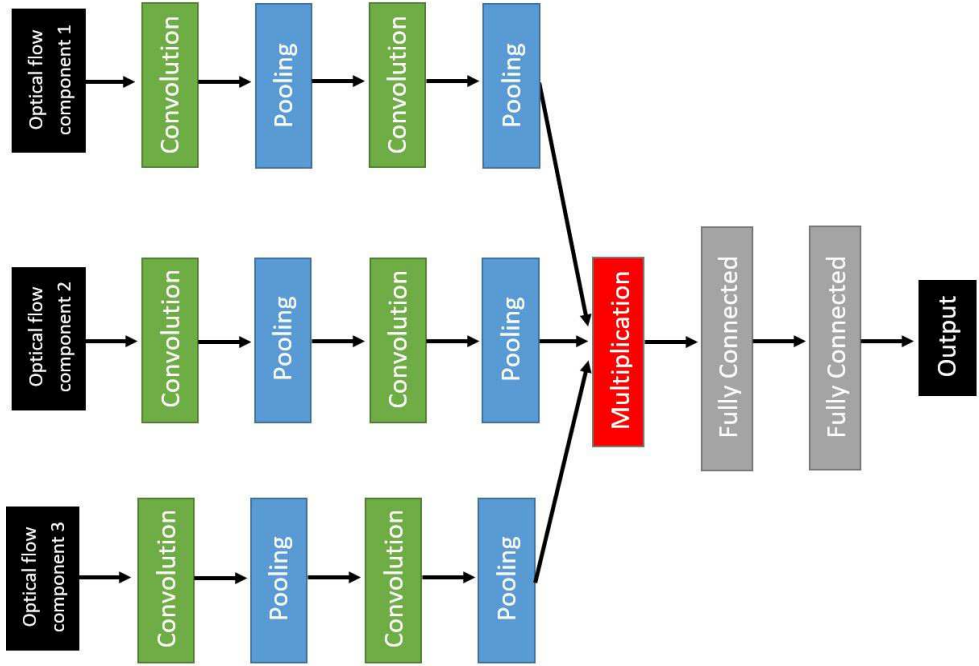}
	\caption{The modified OFF-ApexNet architecture}
	\label{fig:modified_cnn}
\end{figure*}

The modified OFF-ApexNet architecture is illustrated in Figure~\ref{fig:modified_cnn}.

\section{Experiment Setup}
Multiple results analysis on different perspectives are conducted in this paper.
First, we investigate the effect of different types of optical flow method by fixing all the other variables (i.e., the feature extractors and classifier) to constant.
Secondly, the output images generated by AC-GAN and SAGAN are tested to verify its effectiveness in capturing the ME details.  
Thirdly, to examine the recognition performance by varying the input to the modified OFF-ApexNet.
All the analysis are evaluated on a SVM with Leave-One-Subject-Out Cross Validation (LOSOCV) protocol.
Concisely, the video sequence of one subject is treated as the testing data and the rest are categorized as the training data.
This process is repeated for $i$ times, where $i$ is the number of subjects in the database. 
Then, the recognition results for all subjects are averaged to become the final resultant recognition accuracy.
The average accuracy is defined as:

\begin{equation}\label{eq:f-measure}
\text{Accuracy} := \frac{\text{TP+TN}}{\text{TP+TN+FN+FP}}, 
\end{equation}

\noindent where TP, TN, FN and FP are the true positive, true negative, false negative and false positive, respectively.
Due to the imbalance class distribution, F1-score performance metric is computed: 
\begin{equation}\label{eq:f-measure}
\text{F-measure} := 2 \times \frac{\text{Precision} \times \text{Recall}}{\text{Precision + Recall}}, 
\end{equation}
for
\begin{equation}\label{eq:recall}
\text{Recall} := \frac{\text{TP}}{\text{TP + FN}}, 
\end{equation}
and
\begin{equation}\label{eq:precision}
\text{Precision} := \frac{\text{TP}}{\text{TP + FP}}.
\end{equation}

\noindent 
where TP, FN and FP are the true positive, false negative and false positive, respectively.

The details of the experimental setup is elaborated as follows:

\subsection{Optical Flow computation}
\begin{figure*}[t!]
	\centering
	\includegraphics[width=0.7\linewidth]{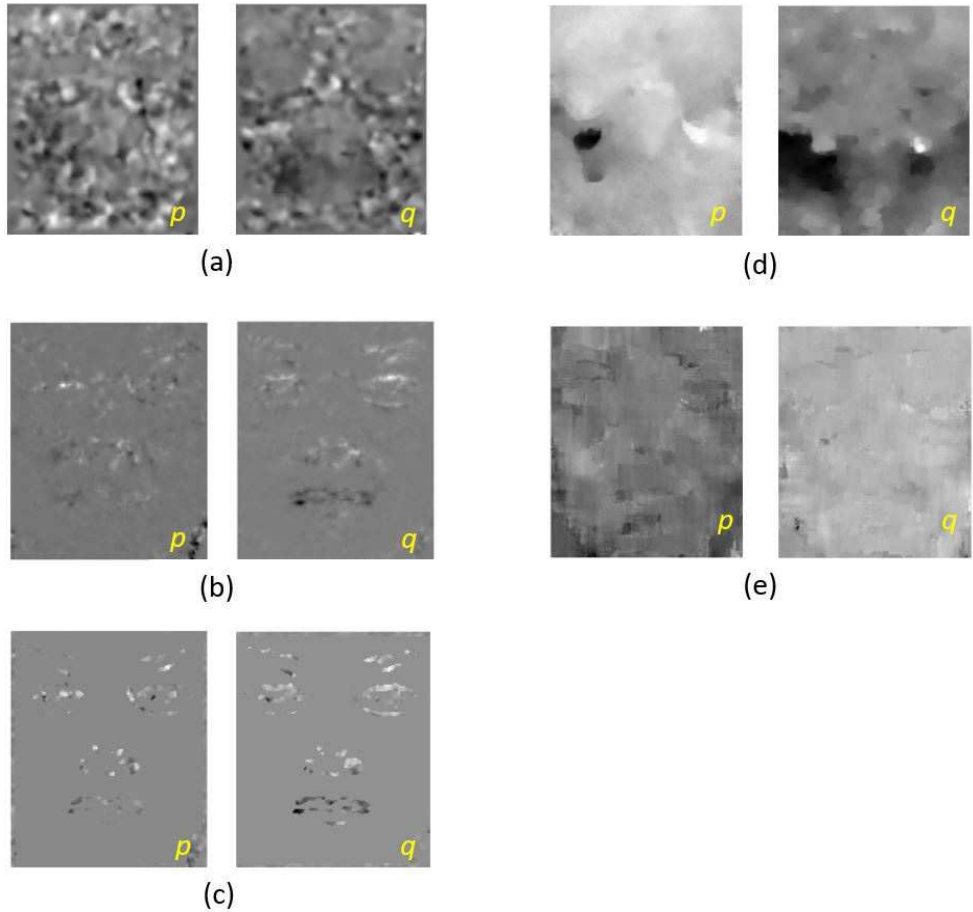}
	\caption{The horizontal and vertical optical flow derived images ($p$ and $q$) of a participant with subtle lip corner puller motion: 
		(a) Farneback;
		(b) Horn \& Schunck;
		(c) Lucas Kanade;
		(d) TVL1, and;
		(e) RLOF
	}
	\label{fig:OF_variation}
\end{figure*}

\begin{figure*}[t!]
	\centering
	\includegraphics[width=0.7\linewidth]{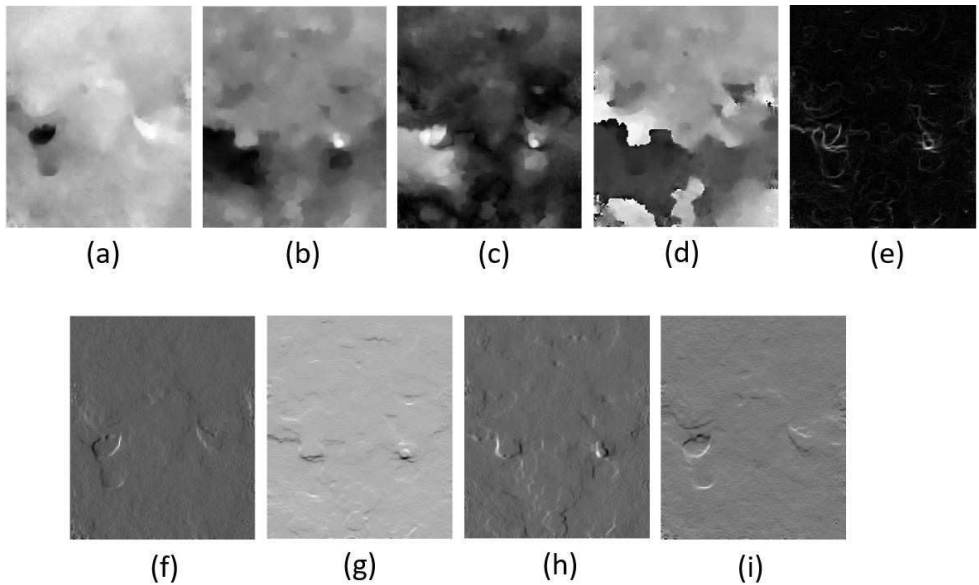}
	\caption{The optical flow derived images of a participant with happiness emotion: (a) $p$; (b) $q$; (c) $\rho$; (d) $\theta$; (e) $\varepsilon$; (f) $\varepsilon_{xx}$; (g) $\varepsilon_{yy}$; (h) $\varepsilon_{yx}$; (i) $\varepsilon_{xy}$}
	\label{fig:tvl1}
\end{figure*}

In our experiments, five types of optical flow methods are computed, namely Farneback, Lucas Kanade and Horn \& Schunck, TVL1 and RLOF.
The example of horizontal and vertical optical flow images generated from the onset and apex frames are shown in Figure~\ref{fig:OF_variation}.
It can be seen that Horn \& Schunck and Lucas Kanade have similar visualization, whereas it is obvious that there is a lip corner movement in TVL1 images. 
On the other hand, the example of the nine optical flow variants generated from the same video sequence is illustrated in Figure~\ref{fig:tvl1}.
Specifically, the participants in this video shows a happiness emotion and the action unit triggered is 12 (i.e., the lip corner puller).
It can be observed that most of the images in Figure~\ref{fig:tvl1} shows obvious movement, or has brighter color at the lip corner region.

We opt to conduct the experiments of this distinct type of optical flow approaches on SMIC database.
Note that, the optical flow are derived from the onset and apex frames using the steps mentioned in Section~\ref{subsec:spot}.
All the 164 video clips from SMIC are considered and three emotion classes involved are surprise, positive and negative. 
Then, the derived $(\rho,\theta,\varepsilon)$ components are served as the input for the Bi-WOOF feature extractor~\cite{liong2018less} to for feature representation and SVM classifier for ME prediction.

\begin{figure}[!tbp]
	\centering
	\begin{minipage}[b]{0.2\textwidth}
		\includegraphics[width=\textwidth]{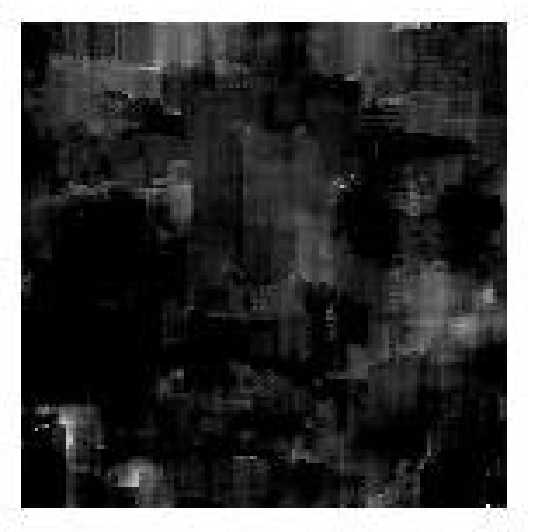}
		\caption{$RLOF_u$}
	\end{minipage}
	\hfill
	\begin{minipage}[b]{0.2\textwidth}
		\includegraphics[width=\textwidth]{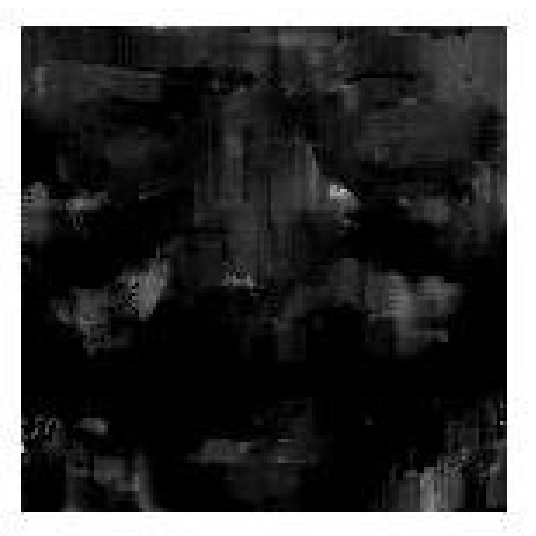}
		\caption{$RLOF_v$}
	\end{minipage}
	\begin{minipage}[b]{0.2\textwidth}
		\includegraphics[width=\textwidth]{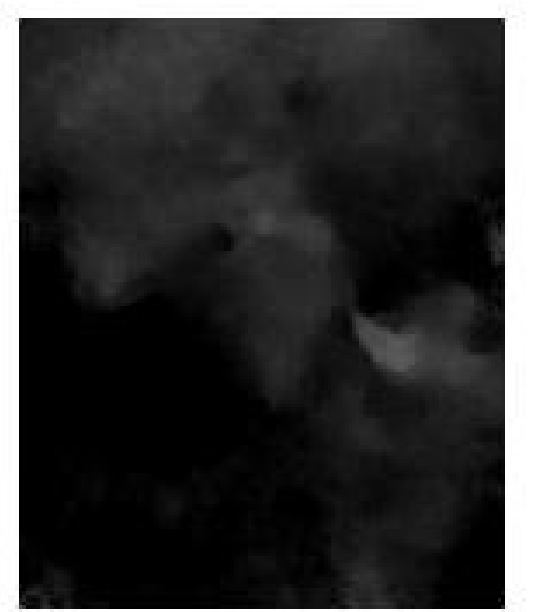}
		\caption{$TVL1_v$}
	\end{minipage}
	\begin{minipage}[b]{0.2\textwidth}
		\includegraphics[width=\textwidth]{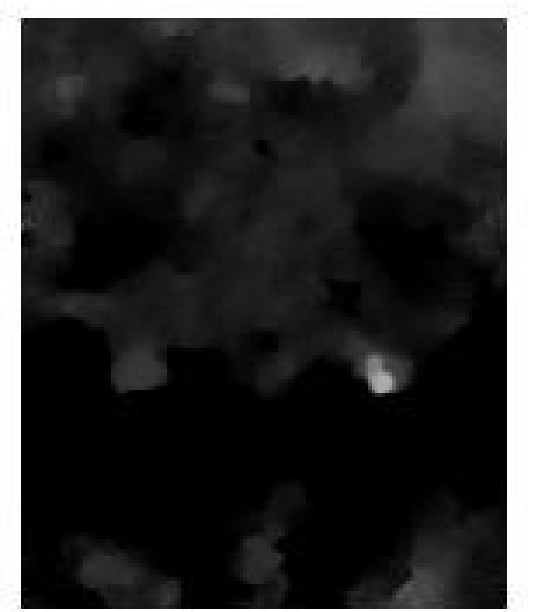}
		\caption{$TVL1_v$}
	\end{minipage}
\end{figure}

\subsection{GAN for Data Augmentation}
\begin{figure}[t!]
	\centering
			\centering
			\includegraphics[width=0.4\textwidth]{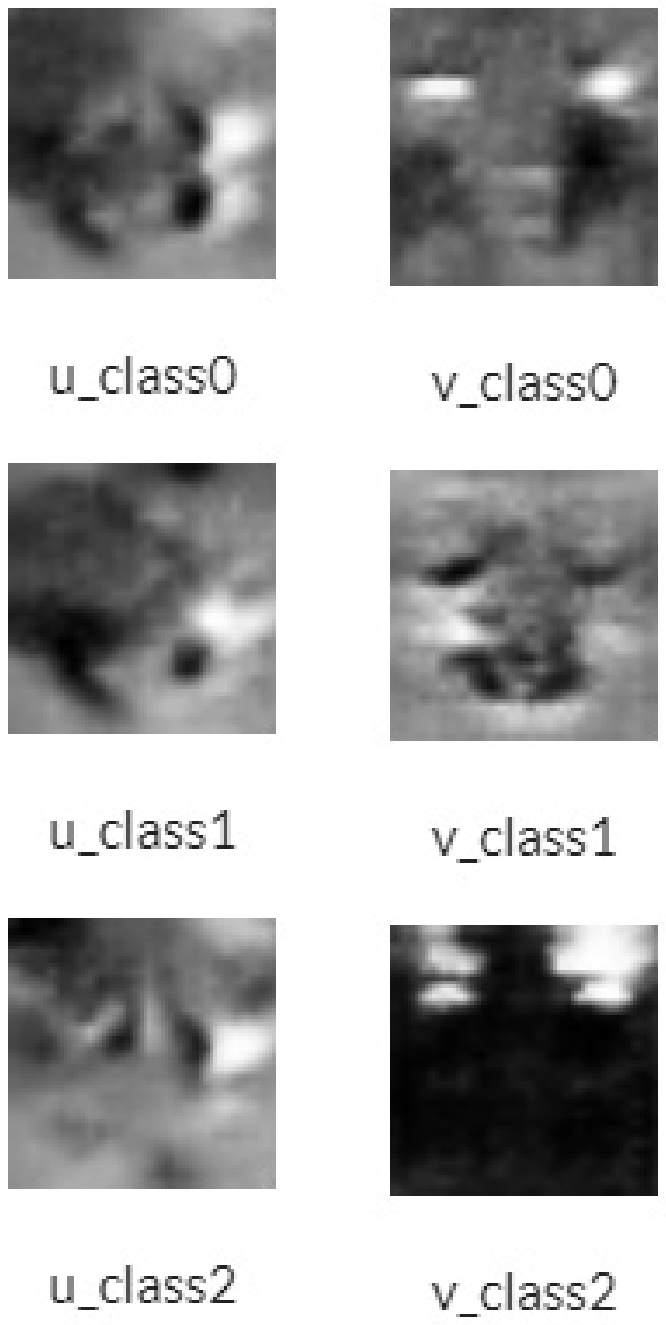}
			\caption{Fake images of AC-GAN}
			\label{fig:acganfake}
\end{figure}

\begin{figure}[t!]
			\centering
			\includegraphics[width=0.4\textwidth]{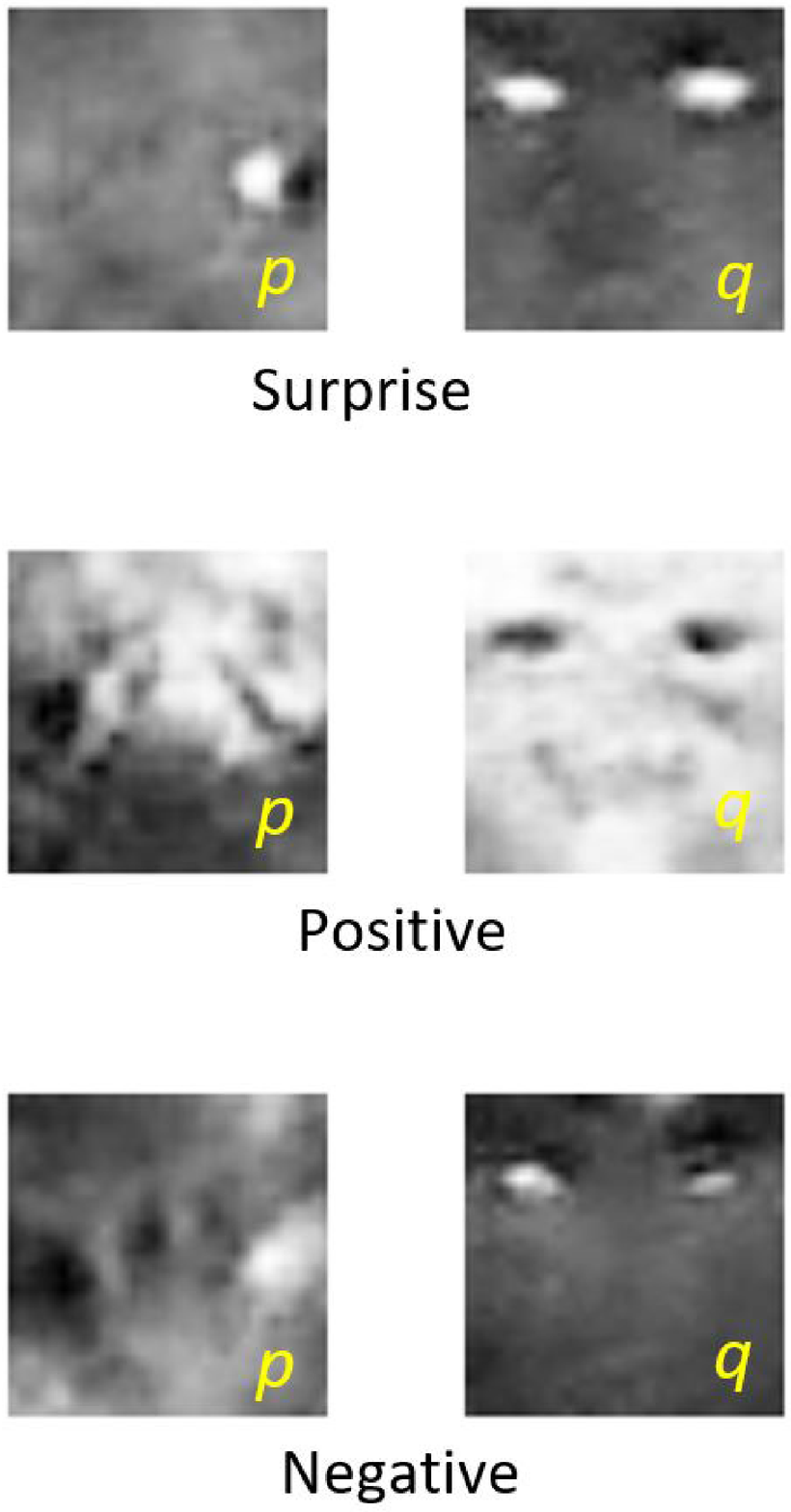}
			\caption{Fake images of SAGAN}
			\label{fig:saganfake}

\end{figure}

AC-GAN and SAGAN are chosen to artificially construct more data sample.
Specifically, only the horizontal and vertical optical flow images are taken into consideration, and they are being trained by the GANs model separately.
The emotion classes and random noise are first fed into the generator.
Next, the same emotion class and the fake images produced by the generator are passed to discriminator to train the GAN model.
Through this, the two individual network structures obtained (i.e., horizontal and vertical optical flow) can be used to create any number of fake images.
The example of the AC-GAN and SAGAN generated images are shown in Figure~\ref{fig:acganfake} and Figure~\ref{fig:saganfake}, with the annotated emotion class and optical flow components.
The goal is to produce the training set to cope with the emotion class imbalance issue.
The experiments of adopting GAN are conducted on SMIC database. 
Then, the Bi-WOOF feature extractor and SVM classifier are adopted to produce the final recognition results.


\subsection{Input to the Modified OFF-ApexNet}
The original OFF-ApexNet was initially evaluated on the composite database which comprised of CASME II, SMIC and SAMM databases.
There are a total 68 subjects with a total of 442 video clips.
The nine optical flow components (i.e., shown in Figure~\ref{fig:tvl1}) are first extracted.
Then they are normalized and reshaped into a $n\times 28\times 28\times 1$ matrix, where $n$ is the number of kernel.
The detailed configuration for the original OFF-ApexNet architecture is listed in Table~\ref{table:CNNsetting}.
The performance result reported range at epoch~=~[100,~5000].
Apart from testing three streams (where $p$ and $q$ images are always remained) for the modified OFF-ApexNet, we also tested single $p$ direction optical flow as the input.
This is to demonstrate that single optical flow component is not sufficient for feature representation and leads to poor recognition performance.

\begin{table}[t!] 
	\begin{center}
		\scriptsize
		\caption{Original OFF-ApexNet configuration~\cite{liong2018off}}
		\label{table:CNNsetting}
		\begin{tabular}{lccccc}
			\noalign{\smallskip}
			\hline
			\noalign{\smallskip}
			Layer
			& Filter size
			& Kernel size
			& Stride
			& Padding
			& Output size \\
			\hline
			\noalign{\smallskip}
			Conv 1
			& 5 $\times$ 5 $\times$ 1 
			& 6
			& [1,1,1,1]
			& Same
			& 28 $\times$ 28 $\times$ 6 \\
			
			\noalign{\smallskip}
			Pool 1
			& 2 $\times$ 2 
			& -
			& [1,2,2,1]
			& Same
			& 14 $\times$ 14 $\times$ 6 \\
			
			\noalign{\smallskip}
			Conv 2
			& 5 $\times$ 5 $\times$ 6 
			& 16
			& [1,1,1,1]
			& Same
			& 14 $\times$ 14 $\times$ 16 \\

			\noalign{\smallskip}
			Pool 2
			& 2 $\times$ 2 
			& -
			& [1,2,2,1]
			& Same
			& 7 $\times$ 7 $\times$ 16 \\
			
			\noalign{\smallskip}
			FC 1
			& -
			& -
			& -
			& -
			& 1024 $\times$ 1 \\
			
			\noalign{\smallskip}
			FC 2
			& -
			& -
			& -
			& -
			& 1024 $\times$ 1 \\
			
			\noalign{\smallskip}
			Output
			& -
			& -
			& -
			& -
			& 3 $\times$ 1 \\
			\hline
			
		\end{tabular}
	\end{center}
\end{table}

\section{Result and discussion }

\subsection{Variation of Optical Flow Method}
The results of varying the optical flow methods using the Bi-WOOF feature extractor and tested on SMIC database is shown in Table~\ref{table:of5}.
It can be seen that TVL1 exhibits the highest accuracy (i.e., 62.20\%) and F1-score (i.e., 61.79\%). 
This type of optical flow method has been adopted in several ME recognition systems, such as~\cite{liong2018off, liong2017micro,gan2018bi}, and promising results are achieved.
Horn \& Schunck and Lucas Kananade methods have the lowest accuracy.
This may due to only a sparse feature sets in the the image are computed.
In contrast, TVL1 considers all the pixels in the frame.
It is worth noted that, for the Farneback method, although the visualization of the $p$ and $q$ images are not showing meaningful ME information (as shown in Figure~\ref{fig:OF_variation} (a)), the recognition results produced are comparable to that of TVL1.

\setlength{\tabcolsep}{5pt}

\begin{table*}[t!]
	\begin{center}
		\caption{Comparison of recognition accuracy (\textit{Acc}  (\%)) and F1-score ($F1$) for different optical flow methods on SMIC database using Bi-WOOF feature extractor.}
		\label{table:of5}
		\begin{tabular}{ccccccccccc}
			\cline{2-11}
			\noalign{\smallskip}
			
			& \multicolumn{2}{c}{Farneback} 
			& \multicolumn{2}{c}{Horn \& Schunck}
			& \multicolumn{2}{c}{Lucas Kanade}
			& \multicolumn{2}{c}{RLOF} 
			& \multicolumn{2}{c}{TVL1}\\
			
			\hline
			\noalign{\smallskip}
			block size
			& Acc
			& F1
			& Acc
			& F1
			& Acc
			& F1
			& Acc
			& F1
			& Acc
			& F1\\
			\hline
			
			\noalign{\smallskip}
			5x5
			& \textbf{59.36}
			& \textbf{.5864}
			& 50.64
			& .4970
			& \textbf{53.04}
			& \textbf{.5284}
			& 51.22
			& .5134
			& \textbf{62.20}
			& \textbf{.6179}
			\\
			
			\noalign{\smallskip}
			6x6    
			& 57.53
			& .5684
			& 50.15
			& .4983
			& 49.45
			& .4390
			& \textbf{54.27}
			& \textbf{.5370}
			& 57.93
			& .5756
			\\
			
			\noalign{\smallskip}
			7x7   
			& 57.83
			& .5726
			& \textbf{51.12}
			& \textbf{0.5082}
			& 51.43
			& 0.5103
			& 51.22
			& .5115
			& 58.54
			& .5742
			\\
			
			\noalign{\smallskip}
			8x8   
			& 58.54
			& .5756
			& 50.83
			& .5067
			& 51.06
			& .5100
			& 51.83
			& .5153
			& 57.32
			& .5672
			\\
			
			\noalign{\smallskip}
			9x9   
			& 57.03
			& .5644
			& 50.42
			& .5018
			& 52.13
			& .5208.
			& 53.05
			& .5302
			& 58.54
			& .5746
			\\
			
			\noalign{\smallskip}
			10x10   
			& 58.23
			& .5772
			& 51.03
			& .4928
			& 52.15
			& .5212
			& 53.05
			& .5312
			& 57.93
			& .5700
			\\
			\hline
			
			\noalign{\smallskip}
			
		\end{tabular}
	\end{center}
\end{table*}

\subsection{Adoption of GAN}

The results including the training set generated by AC-GAN and SAGAN are tabulated in Table~\ref{table:gan}. 
The optical flow images are first reshaped to 28 $\times$ 28 before constructing more images.
Both the recognition results for AC-GAN and SAGAN are presented, with the Bi-WOOF block size ranged from 6 $\times$ 6 to 25 $\times$ 25.
The highest accuracy and F1-score obtained are 61.80\% and 60.23\%, at Bi-WOOF block size~=~15 $\times$ 15.
For SAGAN, the images are first resized to 32 $\times$ 32. 
The highest accuracy and F1-score obtained are 62.20\% and 0.6109, at Bi-WOOF block size~=~8 $\times$ 8.
The experimental results are fluctuating throughout the block size.
It may imply that the GANs are not stable enough and still have rooms for improvement. 

Note that, the the input image size of~\cite{liong2018less} is $170\times140$, whereas we reduced 5 times of the original size, and the recognition performance obtained is comparable to that of reported in~\cite{liong2018less}.
To improve the recognition rate, the image size may be increased to better capture and learn the features.
Besides, the produced training set (including both the real and fake optical flow images), can be evaluated in the deep learning architecture to observe the impact of the size of data sample.

\begin{table}[t!] 
	\begin{center}
		\caption{Comparison of recognition recognition accuracy (\textit{Acc}  (\%)) and F1-score ($F1$) on SMIC database using AC-GAN and SAGAN using Bi-WOOF feature extractor}
		\label{table:gan}
		\begin{tabular}{lccccc}
			\cline{2-5}
			\noalign{\smallskip}
			
			& \multicolumn{2}{c}{AC-GAN} 
			& \multicolumn{2}{c}{SAGAN}\\
			\hline
			\noalign{\smallskip}
			block size
			& Acc
			& F1
			& Acc
			& F1 \\
			\hline
			\noalign{\smallskip}
			6 $\times$ 6
			& 57.01
			& .4963
			& 59.76 
			& .5854 \\
			
			\noalign{\smallskip}
			7 $\times$ 7
			& 48.77 
			& .4575
			& 59.76 
			& .5854 \\
			
			\noalign{\smallskip}
			8 $\times$ 8
			& 50.45
			& .4972
			& \textbf{62.20}
			& \textbf{.6109} \\
			
			\noalign{\smallskip}
			9 $\times$ 9
			& 54.42
			& .4730
			& 61.59
			& .6034 \\
			
			\noalign{\smallskip}
			10 $\times$ 10
			& 52.44
			& .5397
			& 60.37
			& .5940 \\
			
			\noalign{\smallskip}
			11 $\times$ 11
			& 58.84
			& .5194
			& 58.54
			& .5805 \\
			
			\noalign{\smallskip}
			12 $\times$ 12
			& 58.90
			& .4953
			& 60.98
			& .6029 \\
			
			\noalign{\smallskip}
			13 $\times$ 13
			& 50.47
			& .5290
			& 61.59
			& .6104 \\
			
			\noalign{\smallskip}
			14 $\times$ 14
			& 53.26
			& .5442
			& 60.98
			& .6027 \\
			
			\noalign{\smallskip}
			15 $\times$ 15
			& \textbf{61.80}
			& \textbf{.6023}
			& 59.76
			& .5872 \\
			
			\noalign{\smallskip}
			16 $\times$ 16
			& 61.75
			& .6040
			& 59.15
			& .5813 \\
			
			\noalign{\smallskip}
			17 $\times$ 17
			& 58.21
			& .5860
			& 60.37
			& .6021 \\
			
			\noalign{\smallskip}
			18 $\times$ 18
			& 58.82
			& .5939
			& 59.15
			& .5885 \\
			
			\noalign{\smallskip}
			19 $\times$ 19
			& 56.40
			& .5562
			& 57.93
			& .5757 \\
			
			\noalign{\smallskip}
			20 $\times$ 20
			& 56.69
			& .5477
			& 56.71
			& .5605 \\
			
			\noalign{\smallskip}
			21 $\times$ 21
			& 59.35
			& .5584
			& 57.93
			& .5717 \\
			
			\noalign{\smallskip}
			22 $\times$ 22
			& 53.46
			& .5128
			& 57.93
			& .5736 \\
			
			\noalign{\smallskip}
			23 $\times$ 23
			& 53.91
			& .5303
			& 60.37
			& .5944 \\
			
			\noalign{\smallskip}
			24 $\times$ 24
			& 55.54
			& .5444
			& 59.76
			& .5903 \\
			
			\noalign{\smallskip}
			25 $\times$ 25
			& 53.57
			& .5369
			& 59.76
			& .5905 \\
			\hline
			
		\end{tabular}
	\end{center}
\end{table}

\subsection{Modification on OFF-ApexNet}

The results of the original OFF-Apexnet (i.e., $p $ \&  $q$) are reproduced and is tabulated in Table~\ref{table:oriOFF}.
In addition, the performance for the single optical flow component (i.e., $p$) as the input to the modified OFF-ApexNet is reported in the same table for comparison.
It can be seen that for $p $ \&  $q$, the accuracy reached its maximum at epoch~=~ 100 (i.e., 73.01\%) and starts to fluctuate afterwards, whereas the F1-score gradually decreases when the epoch iteration increases.
On the other hand, for $p$, the highest accuracy obtained is 66.21\%, when epoch~=~2000. 
In fact, there is not much accuracy difference within the range of epoch~=~[100,~5000].
It has 7\% and 8\% lower accuracy compared to that of $p $ \&  $q$.
Thus, it can be summarized that using a single optical flow component for feature extraction is not sufficient.
The best accuracy and F1-score reported in~\cite{liong2018off} are 74.60\% and 0.7104, respectively, at epoch~=~3000.
However, when we re-implemented the OFF-ApexNet structure with no changes, both the accuracy and F1-score seems to be lower by 1.5\%.
This amount of deviance is considered acceptable, as the initial parameters (i.e., weights and biases) of the architecture are random, which means that there will be different results for each code execution.

\setlength{\tabcolsep}{5pt}

\begin{table}[t!]
	\begin{center}
		\caption{Comparison of recognition accuracy (\textit{Acc}  (\%)) and F1-score ($F1$) on SMIC, CASME II and SAMM databases using the original OFF-ApexNet and $p$ optical flow component as the input to the modified OFF-ApexNet.}
		\label{table:oriOFF}
		\begin{tabular}{ccccc}
			\cline{2-5}
			\noalign{\smallskip}
			
			& \multicolumn{2}{c}{$p$ \& $q$} 
			& \multicolumn{2}{c}{$p$}\\
			
			\hline
			\noalign{\smallskip}
			Epoch
			& Acc
			& F1
			& Acc
			& F1\\
			\hline
			
			\noalign{\smallskip}
			100
			& \textbf{73.01}
			& \textbf{.6964}
			& 65.53
			& .5971\\
			
			\noalign{\smallskip}
			300
			& 71.20
			& .6805
			& 65.53
			& .6071\\
			
			\noalign{\smallskip}
			500
			& 69.61
			& .6631
			& 65.53
			& .6078\\
			
			\noalign{\smallskip}
			1000
			& 69.84
			& .6667
			& 65.99
			& .6103\\
			
			\noalign{\smallskip}
			2000
			& 70.52
			& .6744
			& \textbf{66.21}
			& \textbf{.6115}\\
			
			\noalign{\smallskip}
			3000
			& 70.29
			& .6712
			& 65.76
			& .6086\\
			
			\noalign{\smallskip}
			4000
			& 70.75
			& .6761
			& 65.99
			& .6113\\
			
			\noalign{\smallskip}
			5000
			& 70.29
			& .6699
			& 65.53
			& .6067\\
			\hline
			
			\noalign{\smallskip}
			
		\end{tabular}
	\end{center}
\end{table}

On another note, the recognition results of the proposed three-stream structure which is modified based on the OFF-ApexNet are shown in Table~\ref{table:multiple}.
Note that the three features obtained after passing to the two convolutional layers and two pooling layers are multiplied, rather than concatenated.
The three input to the architecture are:
(1) $p$ \& $q$ \& $\rho$;
(2) $p$ \& $q$ \& $\theta$;
(3) $p$ \& $q$ \& $\varepsilon$;
(4) $p$ \& $q$ \& $\varepsilon_{ux}$;
(5) $p$ \& $q$ \& $\varepsilon_{uy}$, and;
(6) $p$ \& $q$ \& $\varepsilon_{vx}$;
It is observed that the results in all the multiplication structure, regardless of the input type, are generally lower than that of with two optical flow components as the input (i.e., $p$ \& $q$).
It is obvious that selecting the magnitude component as the third input would be the least preferable option.
For $p \& q \& \varepsilon$, $p \& q \& \varepsilon_{ux}$ and $p \& q \& \varepsilon_{uy}$, the results are comparable, yielding the accuracy of $\sim 71\% $ and the F1-score of $\sim68\%$.

\setlength{\tabcolsep}{5pt}

\begin{table*}[t!]
\begin{center}
\caption{Comparison of recognition accuracy (\textit{Acc}  (\%)) and F1-score ($F1$) on SMIC, CASME II and SAMM databases using the modified OFF-ApexNet.}
\label{table:multiple}
\begin{tabular}{ccccccccccccc}
\cline{2-13}
\noalign{\smallskip}

& \multicolumn{2}{c}{$p$ \& $q$ \& $\rho$} 
& \multicolumn{2}{c}{$p$ \& $q$ \& $\theta$}
& \multicolumn{2}{c}{$p$ \& $q$ \& $\varepsilon$} 
& \multicolumn{2}{c}{$p$ \& $q$ \& $\varepsilon_{ux}$}
& \multicolumn{2}{c}{$p$ \& $q$ \& $\varepsilon_{uy}$} 
& \multicolumn{2}{c}{$p$ \& $q$ \& $\varepsilon_{vx}$} \\

\hline
\noalign{\smallskip}
Epoch
& Acc
& F1
& Acc
& F1
& Acc
& F1
& Acc
& F1
& Acc
& F1
& Acc
& F1\\
\hline

\noalign{\smallskip}
100
& \textbf{67.57}
& .6219
& \textbf{69.16}
& .6270
& 69.16
& .6450
& 68.03
& .5904
& 63.49
& .5264
& 63.49
& .5260\\

\noalign{\smallskip}
300
& 67.12
& \textbf{.6299}
& 68.25
& .6446
& \textbf{71.20}
& \textbf{.6831}
& \textbf{71.88}
& \textbf{.6788}
& \textbf{71.20}
& \textbf{.6764}
& \textbf{66.67}
& \textbf{.6358}\\

\noalign{\smallskip}
600
& 66.21
& .6228
& 68.25
& .6446
& 69.61
& .6621
& 69.16
& .6548
& 68.71
& .6400
& 66.44
& .6281
\\

\noalign{\smallskip}
1000
& 66.43
& .6287
& 66.89
& \textbf{.6599}
& 69.61
& .6599
& 69.16
& .6548
& 68.03
& .6314
& 66.21
& .6265
\\

\noalign{\smallskip}
2000
& 66.43
& .6287
& 66.43
& .6179
& 68.70
& .6498
& 68.93
& .6413
& 66.44
& .6153
& 65.08
& .6137
\\

\noalign{\smallskip}
3000
& 65.53
& .6216
& 66.67
& .6184
& 68.25
& .6454
& 68.48
& .6450
& 66.89
& .6220
& 65.08
& .6121
\\

\noalign{\smallskip}
4000
& 65.76
& .6256
& 66.67
& .6184
& 68.25
& .6472
& 68.02
& .6402
& 67.12
& .6250
& 64.62
& .6077
\\

\noalign{\smallskip}
5000
& 65.99
& .6289
& 66.44
& .6188
& 67.80
& .6434
& 68.25
& .6434
& 66.89
& .6245
& 64.62
& .6077
\\
\hline

\noalign{\smallskip}

\end{tabular}
\end{center}
\end{table*}

\begin{figure}[t!]
\centering
\begin{minipage}[b]{0.4\textwidth}
\includegraphics[width=\textwidth]{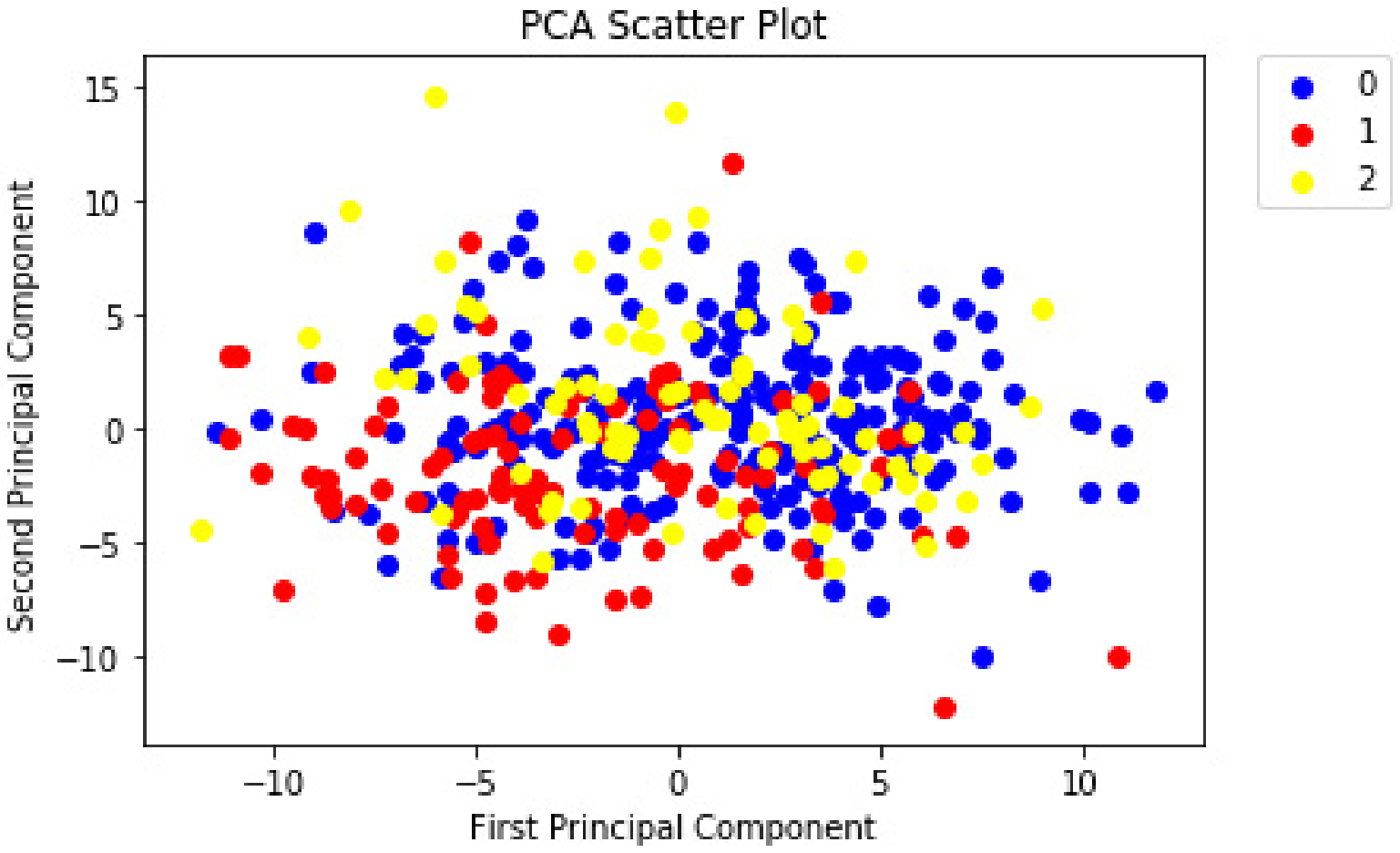}
\caption{The PCA results for $p$ \& $q$ \& $\rho$ as the input data to the modified OFF-ApexNet, at epoch~=~0}
\label{fig:step0_pca}
\end{minipage}
\hfill
\begin{minipage}[b]{0.4\textwidth}
\includegraphics[width=\textwidth]{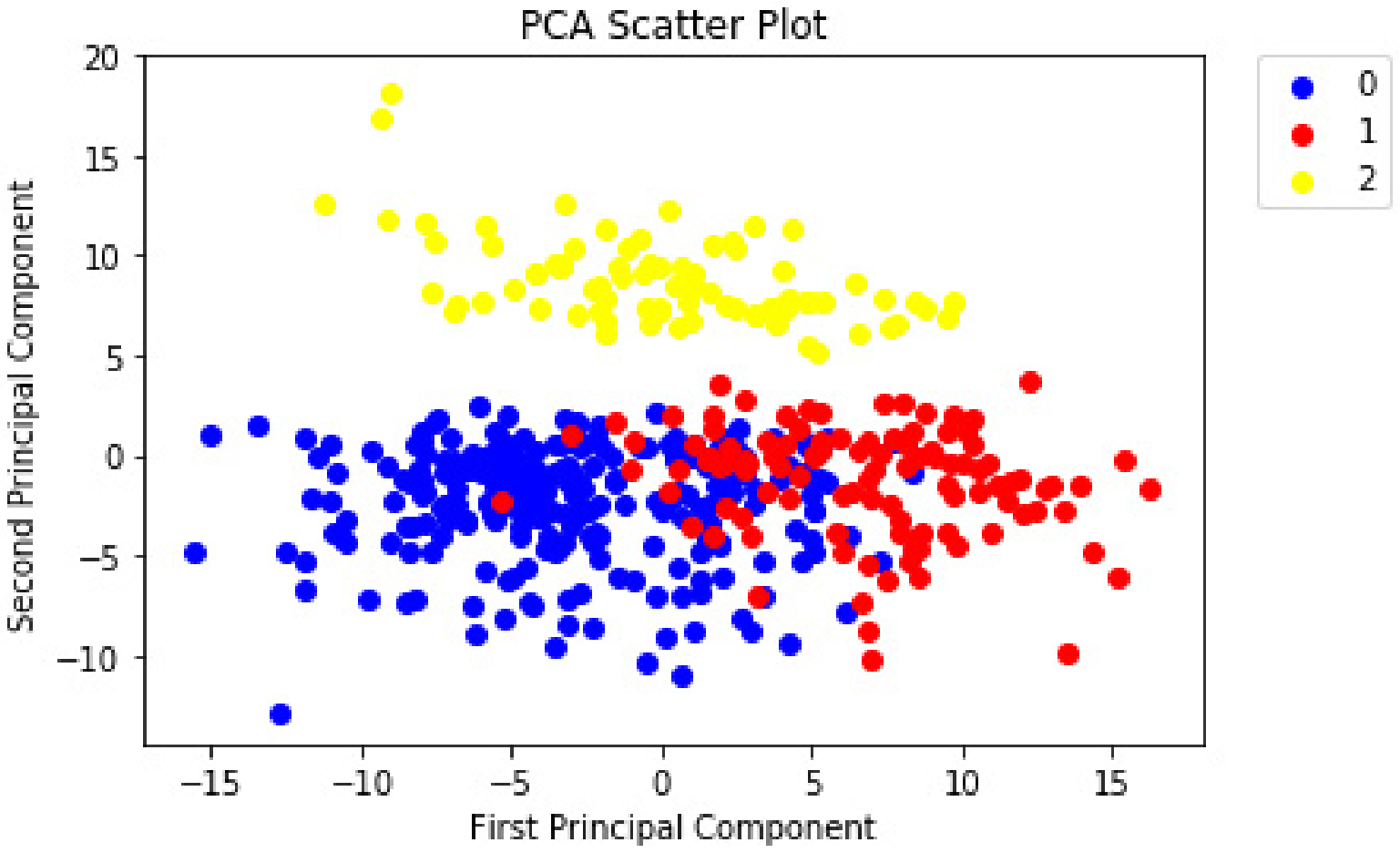}
\caption{The PCA results for $p$ \& $q$ \& $\rho$ as the input data to the modified OFF-ApexNet, at epoch~=~600}
\label{fig:step600_pca}
\end{minipage}
\end{figure}

Furthermore, the results from the principle component analysis (PCA)~\cite{wold1987principal} of certain CNN structures are acquired.
Figure~\ref{fig:step0_pca} and Figure~\ref{fig:step600_pca} are the examples for the $p$ \& $q$ \& $\rho$ c, when the CNN architecture is executed at epoch~=~0 and epoch~=~600, respectively.
The three colors refer to the three emotion classes (i.e., surprise, positive and negative).
As is shown in these two analysis figure, the classifier fails to distinguish the emotion class at the beginning of the training process (i.e., epoch~=~0). 
After the network has been trained for 600 iterations, it shows an obvious trend that one the features scatters for the same class are clustered in a specific regions. 
Therefore, it indicates that this PCA analysis is an optimal tool to visualize the training process to have an insight regarding the features learning behavior.

\section{Conclusion}

This paper analyzes several spatio-temporal features and investigates on two generative adversarial networks (GANs) for micro-expression recognition system.
First, five optical flow variations are computed (i.e., RLOF, TVL1, Farneback, Lucas Kanade and Horn \& Schunck). 
A thorough comparison among these optical flow methods are reported, where we fix all the experiment configurations to observe the effectiveness of each individual method.
Secondly, to deal with the small data size issue and the imbalance of emotion class distribution issues, AC-GAN and SAGAN are employed to generate more artificial micro-expression images.
Thirdly, a slight modification on a state-of-the-art CNN architecture is made.
Comprehensive experiments are conducted and the impact of varying the type of input data are discussed.

For the future works, the GANs generated images with different optical flow methods can be used to enrich the data sample to solve the data imbalance issue.
In addition, the modified OFF-ApexNet can be evaluated by receiving the optical flow components computed by other methods, such as Farneback, RLOF and Lucas Kanade.
Besides, instead of designing the CNN architecture to three parallel stream, it can be extended to more streams to allow automatic high-level feature learning.


\bibliographystyle{ieeetr}

\end{document}